\documentclass[conference]{IEEEtran}
\IEEEoverridecommandlockouts

\usepackage{multirow}
\usepackage{graphicx}
\usepackage{subcaption}

\usepackage{xcolor}

\begin{document}

\title{Impact of Benign Modifications on Discriminative Performance of Deepfake Detectors}

\author{\IEEEauthorblockN{Yuhang Lu, Evgeniy Upenik, and Touradj Ebrahimi}
\IEEEauthorblockA{Multimedia Signal Processing Group (MMSPG)\\
\'Ecole Polytechnique F\'ed\'erale de Lausanne (EPFL)\\
CH-1015 Lausanne, Switzerland\\
Email: firstname.lastname@epfl.ch}
\thanks{The authors acknowledge support from CHIST-ERA project XAIface (CHIST-ERA-19-XAI-011) with funding from the Swiss National Science Foundation (SNSF) under grant number 20CH21\_195532}
}


\maketitle

\begin{abstract}
Deepfakes are becoming increasingly popular in both good faith applications such as in entertainment and maliciously intended manipulations such as in image and video forgery. Primarily motivated by the latter, a large number of deepfake detectors have been proposed recently in order to identify such content. While the performance of such detectors still need further improvements, they are often assessed in simple if not trivial scenarios. In particular, the impact of benign processing operations such as transcoding, denoising, resizing and enhancement are not sufficiently studied. This paper proposes a more rigorous and systematic framework to assess the performance of deepfake detectors in more realistic situations. It quantitatively measures how and to which extent each benign processing approach impacts a state-of-the-art deepfake detection method. By illustrating it in a popular deepfake detector, our benchmark proposes a framework to assess robustness of detectors and provides valuable insights to design more efficient deepfake detectors.


\end{abstract}



\newcommand{\TabResult}{
\begin{table}[ht]
\caption{Detection performance measured by Accuracy and EER, after applying different types of operations to the dataset. GN: Gaussian Noise, GB: Gaussian Blur, GC: Gamma Correction. For bmshj compression, 1 represents lowest compression quality and 8 the highest.}
\label{tab:results}
\resizebox{\linewidth}{!}{%
\begin{tabular}{|clll|}
\hline
\multicolumn{1}{|l}{} & Operations & Accuracy & EER \\ \hline
Raw & - & 80.25 & 25.35 \\ \hline
\multirow{3}{*}{\begin{tabular}[c]{@{}c@{}}Video \\ Compression\end{tabular}} & H265\_c28 & 76.51 & 36.04 \\
 & H264\_c23 & 76.38 & 30.29 \\
 & H264\_c40 & 68.23 & 38.58 \\ \hline
\multirow{3}{*}{\begin{tabular}[c]{@{}c@{}}Image \\ Transcoding\end{tabular}} & JPEG 95 & 78.88 & 26.20 \\
 & JPEG 75 & 76.57 & 28.44 \\
 & JPEG 50 & 72.14 & 31.31 \\ \hline
\multirow{4}{*}{\begin{tabular}[c]{@{}c@{}}Image \\ Smoothing\end{tabular}} & Gaussian Blur ks=3 & 75.50 & 34.24 \\
 & Gaussian Blur ks=5 & 72.60 & 38.36 \\
 & Mean Blur ks=5 & 72.63 & 38.43 \\
 & Median Blur ks=5 & 78.25 & 39.14 \\ \hline
\multirow{2}{*}{\begin{tabular}[c]{@{}c@{}}Additive \\ Noise\end{tabular}} & Gaussian Noise (0,0.01) & 56.61 & 44.17 \\
 & Gaussian Noise (0,0.05) & 27.66 & 50.00 \\ \hline
\multirow{2}{*}{\begin{tabular}[c]{@{}c@{}}Gamma \\ Correction\end{tabular}} & $\gamma$=0.4 & 78.33 & 36.21 \\
 & $\gamma$=2.5 & 79.75 & 28.19 \\ \hline
\multirow{4}{*}{Combination} & GN (0,0.01) + GB ks=5 & 54.33 & 39.94 \\
 & GB ks=5 + GC $\gamma$=0.4 & 74.44 & 45.15 \\
 & GB ks=5 + JPEG 95 & 70.48 & 38.88 \\
 & GC $\gamma$=0.4 + JPEG 95 & 78.07 & 34.46 \\ \hline
\multirow{2}{*}{Resizing} & Linear Interp (scale=1.3) & 83.18 & 22.93 \\
 & Super Res (scale=1.3) & 78.65 & 31.14 \\ \hline
\multirow{5}{*}{\begin{tabular}[c]{@{}c@{}}AI-based\\ Compression\end{tabular}} & hific\_low & 76.93 & 34.40 \\
 & hific\_med & 78.19 & 31.30 \\
 & hific\_high & 77.76 & 30.38 \\
 & bmshj\_1 & 73.28 & 42.91 \\
 & bmshj\_4 & 76.46 & 40.04 \\
 & bmshj\_8 & 78.69 & 33.40 \\ \hline
\end{tabular}%
}
\end{table}
}

\newcommand{\TabResultTwo}{
\begin{table}[ht]
\caption{Detection performance measured by Accuracy and EER, after applying different types of operations to the dataset. GN: Gaussian Noise, GB: Gaussian Blur, GC: Gamma Correction. For bmshj compression, 1 represents lowest compression quality and 8 the highest.}
\label{tab:results}
\resizebox{\linewidth}{!}{%
\begin{tabular}{|cllll|}
\hline
\multicolumn{1}{|l}{} & Operations & Acc & EER & F1-Score\\ \hline
Raw & - & 80.25 & 25.35 & 87.4\\ \hline
\multirow{3}{*}{\begin{tabular}[c]{@{}c@{}}Video \\ Compression\end{tabular}} & H265\_c28 & 76.51 & 36.04 & 85.49\\
 & H264\_c23 & 76.38 & 30.29 & - \\
 & H264\_c40 & 68.23 & 38.58 & 78.77 \\ \hline
\multirow{3}{*}{\begin{tabular}[c]{@{}c@{}}Image \\ Transcoding\end{tabular}} & JPEG 95 & 78.88 & 26.20 & 86.61\\
 & JPEG 75 & 76.57 & 28.44 & 84.65\\
 & JPEG 50 & 72.14 & 31.31 & 80.97\\ \hline
\multirow{4}{*}{\begin{tabular}[c]{@{}c@{}}Image \\ Smoothing\end{tabular}} & Gaussian Blur ks=3 & 75.50 & 34.24 & 84.14\\
 & Gaussian Blur ks=5 & 72.60 & 38.36 & 82.23\\
 & Mean Blur ks=5 & 72.63 & 38.43 & 82.26\\
 & Median Blur ks=5 & 78.25 & 39.14 & 86.86\\ \hline
\multirow{2}{*}{\begin{tabular}[c]{@{}c@{}}Additive \\ Noise\end{tabular}} & Gaussian Noise (0,0.01) & 56.61 & 44.17 & 67.89\\
 & Gaussian Noise (0,0.05) & 27.66 & 50.00 & 22.35\\ \hline
\multirow{2}{*}{\begin{tabular}[c]{@{}c@{}}Gamma \\ Correction\end{tabular}} & $\gamma$=0.4 & 78.33 & 36.21 & 87.09\\
 & $\gamma$=2.5 & 79.75 & 28.19 & 87.26\\ \hline
\multirow{4}{*}{Combination} & GN (0,0.01) + GB ks=5 & 54.33 & 39.94 & 63.89\\
 & GB ks=5 + GC $\gamma$=0.4 & 74.44 & 45.15 & 84.34\\
 & GB ks=5 + JPEG 95 & 70.48 & 38.88 & 80.42\\
 & GC $\gamma$=0.4 + JPEG 95 & 78.07 & 34.46 & 86.78\\ \hline
\multirow{2}{*}{Resizing} & Linear Interp (scale=1.3) & 83.18 & 22.93 & -\\
 & Super Res (scale=1.3) & 78.65 & 31.14 & 86.53\\ \hline
\multirow{5}{*}{\begin{tabular}[c]{@{}c@{}}AI-based\\ Compression\end{tabular}} & hific\_low & 76.93 & 34.40 & 85.66\\
 & hific\_med & 78.19 & 31.30 & 86.36\\
 & hific\_high & 77.76 & 30.38 & 85.95\\
 & bmshj\_1 & 73.28 & 42.91 & 83.48\\
 & bmshj\_4 & 76.46 & 40.04 & 85.65\\
 & bmshj\_8 & 78.69 & 33.40 & 86.79\\ \hline
\end{tabular}%
}
\end{table}
}

\newcommand{\TabResultThree}{
\begin{table}[ht]
\caption{Detection performance measured by Accuracy, AUC and F1-score, after applying different types of operations to the dataset. GN: Gaussian Noise, GB: Gaussian Blur, GC: Gamma Correction. For bmshj compression, 1 represents lowest compression quality and 8 the highest.}
\label{tab:results}
\resizebox{\linewidth}{!}{%
\begin{tabular}{|cllll|}
\hline
\multicolumn{1}{|l}{} & Operations & Acc & AUC & F1-Score\\ \hline
Raw & - & 80.25 & 71.35 & 87.40\\ \hline
\multirow{3}{*}{\begin{tabular}[c]{@{}c@{}}Video \\ Compression\end{tabular}} & H265\_c28 & 76.51 & 61.49 & 85.49\\
 & H264\_c23 & 76.38 & 67.96 & 84.87 \\
 & H264\_c40 & 68.23 & 60.04 & 78.77 \\ \hline
\multirow{3}{*}{\begin{tabular}[c]{@{}c@{}}Image \\ Transcoding\end{tabular}} & JPEG 95 & 78.88 & 70.40 & 86.61\\
 & JPEG 75 & 76.57 & 70.33 & 84.65\\
 & JPEG 50 & 72.14 & 69.17 & 80.97\\ \hline
\multirow{4}{*}{\begin{tabular}[c]{@{}c@{}}Image \\ Smoothing\end{tabular}} & Gaussian Blur ks=3 & 75.50 & 66.91 & 84.14\\
 & Gaussian Blur ks=5 & 72.60 & 62.60 & 82.23\\
 & Mean Blur ks=5 & 72.63 & 62.51 & 82.26\\
 & Median Blur ks=5 & 78.25 & 60.83 & 86.86\\ \hline
\multirow{2}{*}{\begin{tabular}[c]{@{}c@{}}Additive \\ Noise\end{tabular}} & Gaussian Noise (0,0.01) & 56.61 & 55.47 & 67.89\\
 & Gaussian Noise (0,0.05) & 27.66 & 49.58 & 22.35\\ \hline
\multirow{2}{*}{\begin{tabular}[c]{@{}c@{}}Gamma \\ Correction\end{tabular}} & $\gamma$=0.4 & 78.33 & 58.74 & 87.09\\
 & $\gamma$=2.5 & 79.75 & 69.27 & 87.26\\ \hline
\multirow{4}{*}{Combination} & GN (0,0.01) + GB ks=5 & 54.33 & 60.05 & 63.89\\
 & GB ks=5 + GC $\gamma$=0.4 & 74.44 & 57.06 & 84.34\\
 & GB ks=5 + JPEG 95 & 70.48 & 62.54 & 80.42\\
 & GC $\gamma$=0.4 + JPEG 95 & 78.07 & 60.30 & 86.78\\ \hline
\multirow{2}{*}{Resizing} & Linear Interp (scale=1.3) & 83.18 & 75.30 & 89.37\\
 & Super Res (scale=1.3) & 78.65 & 68.02 & 86.53\\ \hline
\multirow{5}{*}{\begin{tabular}[c]{@{}c@{}}AI-based\\ Compression\end{tabular}} & hific\_low & 76.93 & 63.12 & 85.66\\
 & hific\_med & 78.19 & 65.96 & 86.36\\
 & hific\_high & 77.76 & 66.82 & 85.95\\
 & bmshj\_1 & 73.28 & 56.63 & 83.48\\
 & bmshj\_4 & 76.46 & 59.39 & 85.65\\
 & bmshj\_8 & 78.69 & 65.42 & 86.79\\ \hline
\end{tabular}%
}
\end{table}
}

\newcommand\x{0.25\linewidth}
\newcommand\y{\linewidth}
\newcommand{\FigOp}{
\begin{figure}[ht]
\centering
\begin{subfigure}[b]{\x}
  \includegraphics[width=\y]{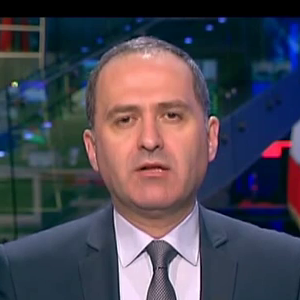}  
  \caption{Raw}
\end{subfigure}%
\hfill
\begin{subfigure}[b]{\x}
  \includegraphics[width=\y]{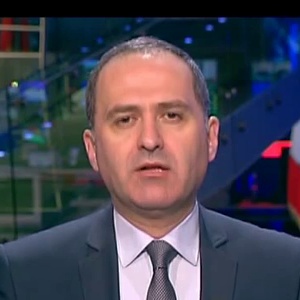}  
  \caption{JPEG 95}
\end{subfigure}%
\hfill
\begin{subfigure}[b]{\x}
  \includegraphics[width=\y]{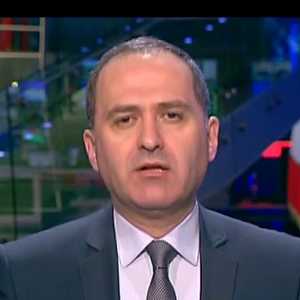}  
  \caption{JPEG 50}
\end{subfigure}%
\hfill
\begin{subfigure}[b]{\x}
  \includegraphics[width=\y]{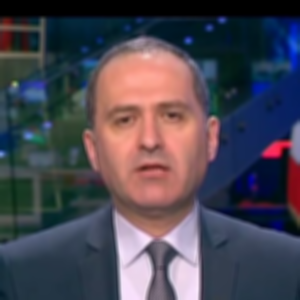}  
  \caption{GB}
\end{subfigure}%
\hfill
\begin{subfigure}[b]{\x}
  \includegraphics[width=\y]{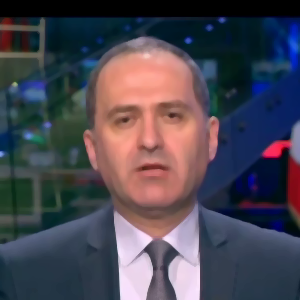}  
  \caption{MB}
\end{subfigure}%
\hfill
\begin{subfigure}[b]{\x}
  \includegraphics[width=\y]{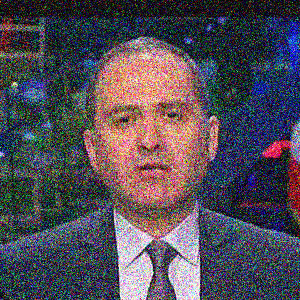}  
  \caption{GN (0,0.05)}
\end{subfigure}%
\hfill
\begin{subfigure}[b]{\x}
  \includegraphics[width=\y]{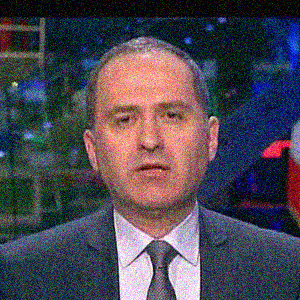}  
  \caption{GN (0,0.01)}
\end{subfigure}%
\hfill
\begin{subfigure}[b]{\x}
  \includegraphics[width=\y]{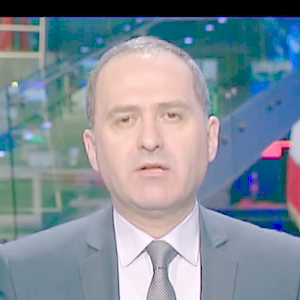}
  \caption{Gamma 0.4}
\end{subfigure}%
\hfill
\begin{subfigure}[b]{\x}
  \includegraphics[width=\y]{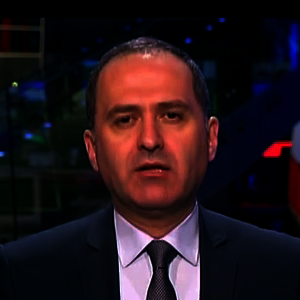}  
  \caption{Gamma 2.5}
\end{subfigure}%
\hfill
\begin{subfigure}[b]{\x}
  \includegraphics[width=\y]{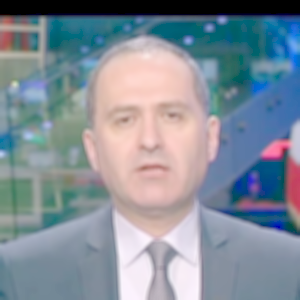}  
  \caption{GB+Gamma}
\end{subfigure}%
\hfill
\begin{subfigure}[b]{\x}
  \includegraphics[width=\y]{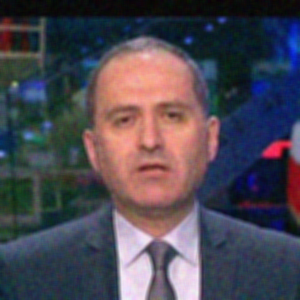}  
  \caption{GN+GB}
\end{subfigure}%
\hfill
\begin{subfigure}[b]{\x}
  \includegraphics[width=\y]{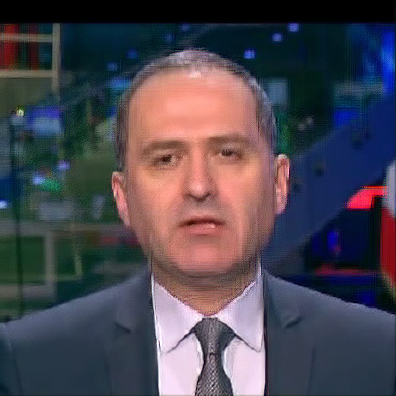}  
  \caption{Super Res}
\end{subfigure}
\hfill
\begin{subfigure}[b]{\x}
  \includegraphics[width=\y]{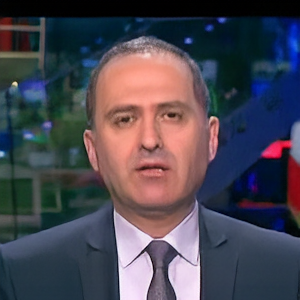}  
  \caption{$hific_{lo}$}
\end{subfigure}%
\hfill
\begin{subfigure}[b]{\x}
  \includegraphics[width=\y]{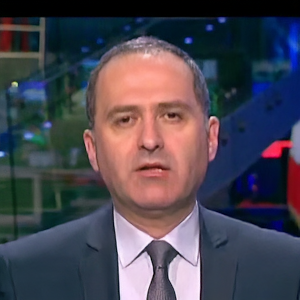}  
  \caption{$hific_{hi}$}
\end{subfigure}%
\hfill
\begin{subfigure}[b]{\x}
  \includegraphics[width=\y]{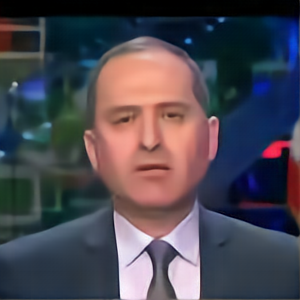}  
  \caption{$bmshj_{lo}$}
\end{subfigure}%
\hfill
\begin{subfigure}[b]{\x}
  \includegraphics[width=\y]{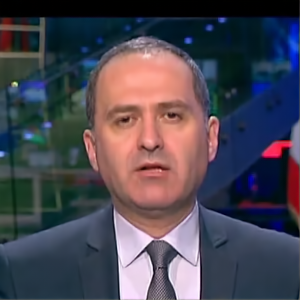}  
  \caption{$bmshj_{hi}$}
\end{subfigure}%
\caption{Example of a typical frame in the dataset, after applying various  operations. GB: Gaussian blur with kernel size 5; MB: Median blur with kernel size 5; GN: Gaussian noise; Super Res: Learning-based super resolution; $hific$ and $bmshj$: Two learning-based image compression algorithms.}
\label{fig:opexample}
\end{figure}
}

\newcommand{\Figjpeg}{
	\begin{figure}[h]
		\centering
		\includegraphics[width=\linewidth]{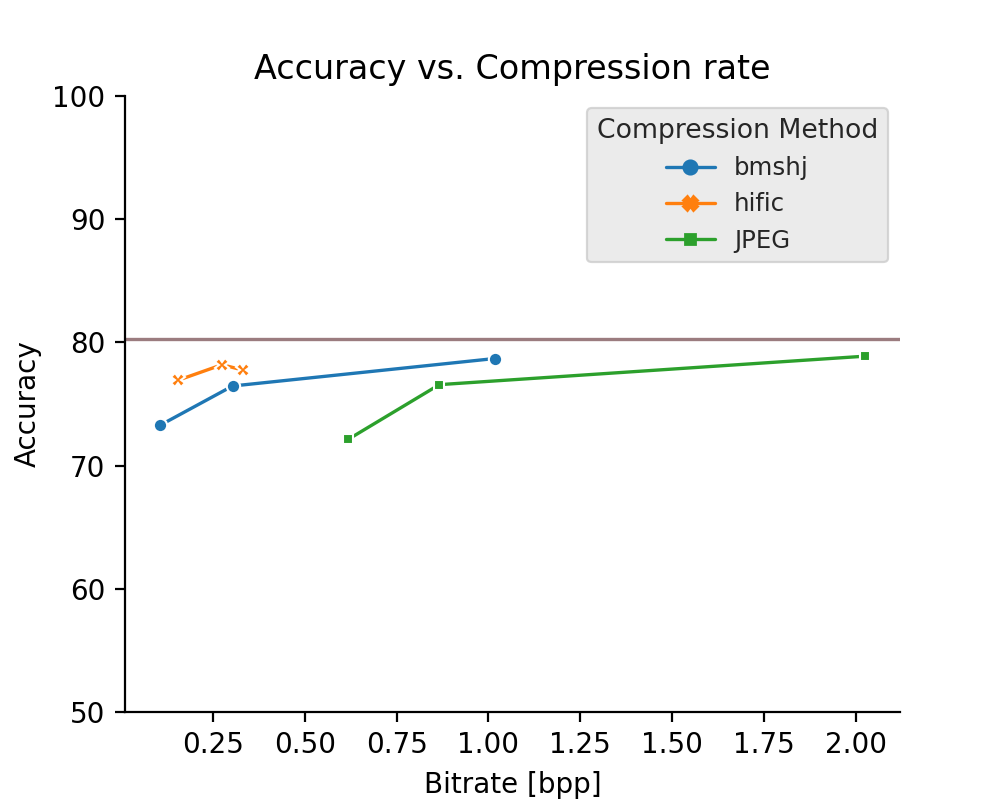}
		\caption{Detection accuracy with respect to three compression algorithms. The horizontal line is the performance on raw data.}
		\label{fig:acc-jpeg}
	\end{figure}
}

\IEEEpeerreviewmaketitle

\section{Introduction}

In recent years, face swapping has become one of the most popular face manipulation techniques due to the great concerns with the deepfake generation. The deep learning-based tools and open source software have simplified the creation of such manipulated contents, posing serious public concerns. It is crucial to develop deepfake detection systems that can automatically and effectively identify manipulated videos and images.
However, while the current state-of-the-art detectors achieve  promising results on target databases, they are trained and tested in rather simple scenarios and often show poor performance in post-processed contents. In realistic situations, videos and images on the Internet and social networks constantly undergo benign processing to ease transmission, including but not limited to compression, denoising, enhancement and resizing. A deepfake detector should be robust to such modifications. Moreover, hackers can also make use of such drawbacks and easily fool the forensic detection  by post-processing.

Some recent works \cite{rossler_faceforensics_2019,jiang_deeperforensics-10_2020,dolhansky_deepfake_2020} have considered this problem when creating the database. For example, FaceForensics++ database~\cite{rossler_faceforensics_2019} compresses raw video footage based on H.264  at two different quality factors. Jiang et al. \cite{jiang_deeperforensics-10_2020} extensively applied real-world perturbations to enlarge the dataset and meanwhile obtain a more challenging and diverse benchmark. The most recent large scale deepfake database DFDC \cite{dolhansky_deepfake_2020} introduces various overarching types of augmentations. Despite their advancement, neither these database providers nor researchers working in this area, systematically compare and summarize the impact of perturbations by means of post-processing. A more comprehensive and systematic approach with in-depth analysis of how the benign processing impacts the detection performance is desired.   

In this work, an extensive evaluation of the performance of a popular deepfake detection approach towards different modifications is presented as a first step towards explainable artificial intelligence behind deepfake generation and detection. We conduct our study on FaceForensics++ dataset. To generate a broad range of variations on the test data, popular post-processing operations such as video compression, image transcoding, smoothing, and enhancement re applied to deepfakes. In addition, we also modify the image with additive Gaussian noise. As learning-based image processing algorithms have become very popular, we also include learning-based compression and super-resolution methods in our framework. While a performance degradation is expected after applying one of these processing methods, our framework quantifies the impact of each type of processing by showing to which extent the detector under analysis is sensitive to such modifications.

The paper is organized as follows: Section 2 provides a short overview of deepfake creation, detection and popular databases.
Section 3 details the proposed assessment framework. Results are discussed in Section 4 and  Section 5 draws conclusions.

\section{Related Work}

The related work can be organized in three categories, namely, deepfake generation, deepfake detection, and datasets. For a broader review readers are refer to \cite{tolosana_deepfakes_2020}.

\subsection{Deepfake Generation}

Even though studies on face manipulation appeared as early as in the first decade of this century \cite{bitouk_face_2008}, the visual quality of the produced deepfakes only reached acceptable levels with introduction of deep convolutional (CNN) and generative adversarial (GAN) neural networks \cite{korshunova_fast_2017}.

Currently, among the successful deepfake generation methods one can note, particularly: \textit{face swapping}, \textit{lip syncing}, and \textit{motion models}. Face swapping methods typically substitute the face of a subject with that of a target~\cite{perov_deepfacelab_2020}. Lip syncing only changes the lip movements of a face in a way that the generated movements correspond to a particular speech \cite{prajwal_towards_2019,prajwal_lip_2020}. Motion models take a still image of a face as an input and animate it according to a reference from another face \cite{siarohin_first_2019}.

\subsection{Deepfake Detection}

One of the first methods for deepfake detection was proposed by Zhou et al. in \cite{zhou_two-stream_2017}. They use two parallel networks: one detects image tampering and another performs steganalysis to account for camera processing and local noise characteristics.
Afchar et al. \cite{afchar_mesonet_2018} propose an approach that is claimed to be robust to compression artifacts, thanks to excluding explicit noise from consideration.

In \cite{nguyen_multi-task_2019}, Nguyen et al. proposed a convolutional neural network with a Y-shaped autoencoder that can quickly adapt to deal with unseen attacks by using only a few samples for fine-tuning. In \cite{nguyen_use_2019}, the same authors introduce a capsule network that can detect various types of attacks, ranging from presentation attacks using printed images to attacks using fakes created by deep learning.

\subsection{Available Datasets}

Various datasets have been proposed and used by the research community for training deepfake detectiors. Korshunov and Marcel \cite{korshunov_deepfakes_2018} were among the first to release a public dataset of deepfake videos in 2018 that was created using a GAN-based
face-swapping algorithm. A bigger and more popular dataset \textit{FaceForensics++} was released a year later by Rossler et al. \cite{rossler_faceforensics_2019}. This dataset consists of more than 1.8 million images that were face-manipulated as well as their ground truth. In early 2020, major companies partnered with academia to organize the \textit{Deepfake Detection Challenge}\footnote{https://www.kaggle.com/c/deepfake-detection-challenge/} that led to the creation of a dataset \cite{dolhansky_deepfake_2020}. Finally, Li et al.~\cite{li_celeb-df_2020} created the \textit{Celeb-DF} dataset by face-swapping videos using synthesized images of celebrities.

\section{Assessment Framework}
\subsection{Assessment Database}

FaceForensics++ \cite{rossler_faceforensics_2019} was used as the main database for our assessment framework. FaceForensics++ is one of the most popular large-scale databases in the media forensic detection field. It is an extension of the original FaceForensics database \cite{rossler2018faceforensics}, which was initially focused on facial reenactment through Face2Face \cite{thies2016face2face} manipulation. FaceForensics++ contains 1000 real videos collected from YouTube and 4000 fake videos generated using both classical computer graphics approaches, i.e. FaceSwap\footnote{https://github.com/MarekKowalski/FaceSwap/} and Face2Face \cite{thies2016face2face}, and learning-based approaches, i.e. Deepfake\footnote{https://github.com/deepfakes/faceswap}, NeuralTextures \cite{thies2019deferred}. Every two pristine videos were paired up and swapped faces using the four manipulation methods respectively.

For a realistic setting, all videos are compressed using the H.264 codec, which is common in real-world social media and websites, in particular. In addition to raw (no compression), high quality, and low quality compressed videos were generated using a quantization parameter equal to 23 and 40. To our knowledge, FaceForensics++ was one of the few databases that considers different levels of video quality.

The dataset was split into 720 videos for training, 140 for validation, and 140 for testing. To ease the assessment and to establish a uniform evaluation framework, the default split up was performed and focused on the 140 testing videos. In addition, the first 10 frames of each testing video was selected and cropped around facial areas. The cropped images were then resized to the same dimensions and saved in PNG format.

\subsection{Detection Method}
The publicly available detector called  Capsule-Forensics  and proposed by Nguyen et al. \cite{nguyen_use_2019}, was adopted.\footnote{https://github.com/nii-yamagishilab/Capsule-Forensics-v2}.

The Capsule-Forensics approach was selected in our study for several reasons. It combines the traditional CNN and the Capsule Networks, which require fewer parameters with reasonable computational resources when compared to approaches that are based on CNN only, while achieving similar detection performance. The lightweight architecture is more time efficient during evaluation, making it a suitable candidate for our exhaustive assessment framework. Moreover, this approach demonstrates high accuracy on FaceForensics++ dataset and provides a reasonable baseline for the proposed framework.

Furthermore, the original authors used VGG19 \cite{simonyan2014very} pretrained on ImageNet \cite{ILSVRC15} as feature extractor, followed by several primary capsules and two output capsules (‘real’ and ‘fake’). The input images were pre-processed by cropping the face area using a face detection algorithm, then resized to 300x300, which they claim to be large enough to provide reliable results. In addition,  two regularizations were introduced, to reduce overfitting, by adding random noise and a dropout operation during training. In this paper, the pretrained model released by the authors was directly applied in our experiments.





\subsection{Processing Operations}
To create a comprehensive and systematic assessment framework,  consider multiple benign  operations based on both conventional and deep learning-based approaches were considered. As most of the current forgery detection models are evaluated in trivial settings, the main objective here is to see how commonly used operations in real situations affect the state-of-the-art detectors. The details of all  operations used in evaluations are described below with the illustration of a typical example provided in Figure~\ref{fig:opexample}.
\FigOp

Lossy compression refers to the class of data encoding methods that remove unnecessary or less important information and only uses partial data to represent the content. These techniques are used to reduce data size for efficient storage and transmission content and are widely applied to images and videos on social networks and websites. It is interesting to observe how the distortion or artifacts by lossy data compression would affect the performance of a deepfake detector. The FaceForensics++ database provides two compressed versions based on H.264. An additional version was created using  libx265 with the parameter 28, that typically corresponds, visually speaking, to libx264  at CRF 23. For images, JPEG compression with three quality factors was applied to reveal the impact of conventional image compression. 
In image processing, blurring, also known as smoothing, is a widely used operation, typically to reduce noise which at the same time results in reduction of details.  Raw frames were processed with different filters such as Gaussian, and other popular blur operations with varying kernel sizes. A contrasting operation is the addition a Gaussian noise to frames in the dataset, which often occurs during data acquisition by image sensors. We also considered gamma correction as an image enhancement technique, which corrects the brightness of a frame by using a non-linear transformation between the input values and the mapped output. Moreover, a mixture of two or three operations was also considered, such as combining Gaussian noise and Gaussian blur techniques, making the database to better reflect more complex real-world scenarios. 

Recently, thanks to breakthroughs in both hardware and machine learning, a large number of image processing techniques have been proposed based on deep learning, including image upsampling, image compression, image enhancement and so on. For example, Fabian Mentzer et al. \cite{mentzer2020high} directly optimize the neural network for a better rate-distortion trade-off and leverage GANs to prevent compression artifacts, which yields reconstructions with higher fidelity. With promising outcomes from these methods, it is also important to analyze the impact of  learning-based processing on deepfake detection. Therefore, in addition to the conventional approaches, a data-driven super resolution technique \cite{wang2018esrgan} and two generative lossy compression algorithms \cite{mentzer2020high,balle2018variational} were evaluated, in comparison to linear interpolation for resizing and the classical JPEG, H.264 and H.265 compression.

\subsection{Implementation Details}
The entire assessment framework is implemented in PyTorch with Python. The checkpoint used were trained using the Adam optimizer \cite{kingma2014adam} with $\beta1$ = 0.9, $\beta2$ = 0.999, and a learning rate of 5 × $10^{-4}$. A total of 720 pristine videos and 2880 deepfakes were used for training. The first 100 frames of each training video was chosen to train the model for 25 epochs. Notably, the detector was trained on purely unprocessed genuine and fake contents.

The preprocessing step is comprised of video compression, image processing, face detection and resizing.
The proposed framework uses the entire test set of FaceForensics++ database, specifically 140 real videos and the same amount of fake videos for each of the four manipulations. We then extract the first 10 frames of each video and use a face detection to crop and align the frames. Afterwards, multiple benign processing operations are applied independently and constitute the dataset for different assessments.

For a fair comparison with the results obtained by a baseline configuration, the binary detection accuracy is used to rate the performance. Considering the imbalanced test data and to better estimate the system robustness, Area under the Curve (AUC) of Receiver Operating Characteristic Curve (ROC) and F1-score as the metrics are also added to experimental results.



\section{Experimental Results and Discussion}
The performance of the state-of-the-art forensic detection methods towards multiple benign modifications has been assessed. The findings show that preprocessing with even benign operations can have a negative impact on the robustness of a deepfake detector and generally results in noticeable decline of detection performance. In Table~\ref{tab:results}, the accuracy, AUC and F1-score are reported for the model Capsule-Forensics \cite{nguyen_use_2019}.

\TabResultThree
With the exception of resizing, the best performance is achieved on raw videos with an accuracy of 80.25\%. The performance is somewhat lower than the reported score in the original paper because of differences in face detection and training, among others. However, this does not impact the conclusions as the goal is to assess impact on performance after applying various benign operations.

\textit{Performance w.r.t data compression and transcoding} It is expected that the performance of the detector degrades as the raw videos are compressed with lower quality (larger coding parameters). The model achieves 76.38\% accuracy when the test videos are compressed based on libx264 encoder at CRF 23 and it drops to 68.23\% when at CRF 40. It is interesting to observe that the detector's performance on libx265 encoded video at CRF 28 is similar as on H.264 codec at CRF 23, while it results in about half the bitrate. Similar reduction trends can be observed in  JPEG as well as in learning-based compression. The  performance only decreases by 1.37\% on JPEG compressed images with high quality (95\%), while the performance drops quickly as we lower the image quality. The trend is shown in Figure~\ref{fig:acc-jpeg}. For both learning-based lossy compression algorithms, we evaluate the lowest, medium, and highest quality levels. It is remarkable that the high quality learning-based image compression results in comparable performance as high image quality JPEG compression.

\textit{Performance w.r.t image smoothing} Three blurring techniques are evaluated in our framework and all of them lead to performance degradation. The median filter brings less negative impact and reduces the accuracy by only 2\%, while the Gaussian  and mean filters reduce by 7.65\% and 7.62\%. 

\textit{Performance w.r.t noise} Gaussian noise has a significantly negative impact on  performance when compared to all other operations. The detector performs only with 56.61\% accuracy when test images are corrupted by additive Gaussian noise with mean 0 and variance 0.01, nearly close to random guesses. The results are even worse after increasing the variance, potentially posing great threat to current deepfake detectors. We evaluate a mixed operation that adds Gaussian noise first, followed by a denoising Gaussian filter. Interestingly, the additional denoising operation makes the result even worse. Overall, the combination of two or three processing operations degrade the detection accuracy further when compared to only considering one operation independently.

\textit{Performance w.r.t resizing} The test image were resized with a scale of 1.3 which unexpectedly brings improvements detection accuracy, which is consistent with the observations of the original author of Capsule-Forensics. On the contrary, the performance on re-scaled images through learning-based super resolution technique results in slightly decreased performance, possibly due to the accompanying artifacts. However, more experiments are expected to investigate the reasons behind.

\textit{Performance w.r.t image enhancement} It was also considered to change the contrast by underexposing and overexposing frames using gamma operations with ${\gamma=0.4}$ and ${\gamma=2.5}$. As shown from Table~\ref{tab:results}, both operations decrease the performance by a rather small margin and the detector achieves accuracy of 78.33\% and 79.75\% respectively.


\Figjpeg


\section{Conclusions}

In this work, a rigorous and systematic assessment framework for deepfake detectors in realistic situations was presented. The impact of both conventional and learning-based benign modifications on deepfake detectors were taken into account. The usefulness of the proposed framework has been illustrated by conducting an extensive evaluation on Capsule-Forensics approach with processed FF++ database. The benchmark identifies and quantifies the different influences that each benign modification can bring to a deepfake detector. The proposed framework can be adapted to a wider variety of detectors and recognition tasks, and applied to different datasets, in order to obtain more comprehensive and insightful conclusions to other types of detectors and recognition algorithms. Such analysis can become useful for design of more efficient and robust detectors and recognition tasks.






\bibliographystyle{IEEEtran}
\bibliography{refs}

\end{document}